\documentclass{article}

\usepackage[preprint]{neurips_2025}

\usepackage[utf8]{inputenc} 
\usepackage[T1]{fontenc}    
\usepackage{hyperref}       
\usepackage{url}            
\usepackage{booktabs}       
\usepackage{amsfonts}       
\usepackage{nicefrac}       
\usepackage{microtype}      
\usepackage{xcolor}         

\usepackage{tikz}
\usetikzlibrary{shapes.geometric, arrows, fit}
\tikzstyle{box} = [rectangle, text centered, draw=black, line width=1.5pt]
\tikzstyle{arrow} = [thick, ->, >=stealth, align=center, line width=1.5pt]

\usepackage{subcaption}
\usepackage{caption}
\usepackage{mathtools}
\usepackage{pdfpages}
\usepackage{algorithm}
\usepackage{algpseudocode}

\title{GIFT: Global stabilisation via Intrinsic Fine Tuning}

\author{%
  Rory Young\thanks{Corresponding email: R.Young.4@research.gla.ac.uk} \hspace{2em} Nicolas Pugeault \\
  School of Computing Science\\
  University of Glasgow\\
}

\begin{document}

\maketitle

\begin{abstract}
    Deep reinforcement learning policies achieve strong performance in complex continuous control environments with nonlinear contact forces. However, these policies often produce chaotic state dynamics, with trivially small changes to the initial conditions significantly impacting the long-term behaviour of the control system. This high sensitivity to initial conditions limits the application of Deep RL to real-world control systems where performance and stability guarantees are often required. To address this issue, we propose \textbf{G}lobal stabilisation via \textbf{I}ntrinsic \textbf{F}ine \textbf{T}uning (GIFT), a general-purpose training framework which directly optimises the global stability of existing high-performing deep RL policies using a custom reward function. We demonstrate that GIFT increase the stability of the control interaction while maintaining comparable task performance, thereby improving the suitability of deep RL policies for real-world control systems.

\end{abstract}

\section{Introduction}
Deep Reinforcement Learning (RL) \citep{sutton_1998_reinforcement} has emerged as a powerful approach for learning control strategies in continuous domains \citep{mnih_2015_humanlevel, lillicrap_2019_continuous, silver_2016_mastering, silver_2017_mastering_a, silver_2017_mastering_b}. However, while deep RL has achieved remarkable success in simulated benchmark tasks, its deployment in real-world applications remains limited \citep{Degrave2022Magnetic}. While several reasons for this dichotomy exist \citep{dulacarnold_2021_an}, one critical limitation is that learned policies often induce unstable or chaotic feedback loops \citep{Young2024Enhancing}. When this occurs, small perturbations in initial conditions can lead to significant deviations in long-term outcomes \citep{DeterministicNonperiodicFlow, devaney_2003_an}. This high sensitivity to initial conditions limits the application of Deep RL to real-world systems, as these often require guarantees of long-term performance and stability \citep{dulacarnold_2021_an}.

One factor which contributes to this instability is the specification of the reward function. Control systems governed by identical transition dynamics can exhibit significantly different stability characteristics depending on the reward signal used to train the policy. This variation arises because the reward function implicitly determines which state space dimensions should be constrained by the policy. In particular, stable converging behaviour only emerges when the reward function constrains all the dimensions of the system's state \citep{Young2024Enhancing}. In contrast, reward functions that only constrain a subset of the state dimensions can produce policies agnostic to unconstrained dimensions, and this underspecificity can lead to chaotic state dynamics.

To address this issue, we propose \textbf{G}lobal stabilisation via \textbf{I}ntrinsic \textbf{F}ine \textbf{T}uning (GIFT). This lightweight framework directly optimises the global stability properties of existing deep RL policies by fine-tuning them in a Stabilising Markov Decision Process (S-MDP). The S-MDP retains the same transition dynamics, state space, action space, and observation function as the original MDP \citep{sutton_1991_dyna} but it leverages a uniquely generated reward function that incentivises global stability. By constructing an auxiliary reward function that explicitly penalises divergent trajectories, GIFT encourages the policy to regulate all dimensions of the state space that could otherwise lead to instability if left unconstrained. 

We demonstrate that applying GIFT to state-of-the-art deep RL policies in continuous control domains consistently produces significant improvements in stability. Specifically, the resulting policies exhibit Maximal Lyapunov Exponent (MLE) \citep{lyapunov_1992_the} that is an order of magnitude lower than those of the original policies. Moreover, this improvement in stability does not compromise task performance. This makes GIFT a practical and broadly applicable tool for stabilising RL agents in continuous control domains.
\section{Background}
\subsection{Reinforcement Learning}
In general, any contol interaction can be represented by a Markov Decision Process (MDP) with state space $(\mathcal{S} \subseteq \mathbb{R}^{n})$, observation space $(\mathcal{O}\subseteq\mathbb{R}^k)$, action space $(\mathcal{A} \subseteq \mathbb{R}^{m})$, scalar reward function $(r: \mathcal{S} \times \mathcal{A} \rightarrow \mathbb{R})$, state transition function $(f: \mathcal{S} \times \mathcal{A} \rightarrow \mathcal{S})$, observation function $(O: \mathcal{S} \rightarrow \mathcal{O})$, and initial state distribution $(\rho_0 \subseteq \mathcal{S})$. In this framework, the objective of a controller $(\pi_\theta: \mathcal{O} \rightarrow \mathcal{A})$ is to maximise the total sum of rewards over a fixed horizon $(\sum r_t)$. Reinforcement Learning agents generate such a controller by optimising the $\theta$ to maximise the expected return  (Equation~\ref{eq:objective_function}) for a given discount factor $\gamma \in [0,1)$.

\begin{equation}
    \label{eq:objective_function}
    J(\theta) = \mathop{\mathbb{E}}_{s_0 \sim \rho_0} \left [ \left. \sum_{t=0}^{\infty} \gamma^t \times r(s_t, a_t) ~\right\vert~ {s_{t+1}=f(s_t,a_t),~a_t=\pi_\theta(O(s_t))} \right ]
\end{equation}

\subsection{Stable Reinforcement Learning}
Control policies are said to be stable if they can guide a control system to an equilibrium state or state trajectory after a perturbation. This is a critical property for control systems deployed in real-world environments, as disturbances to the system are inevitable. Small deviations from the expected behaviour can accumulate over time, leading to unsafe or undesirable outcomes. A stable policy mitigates this risk by ensuring that the system's response remains bounded and predictable. However, unlike classical controllers, deep RL policies do not inherently incorporate any structure that ensures such behaviour.

Recently, a growing body of work has focused on addressing the stability issue with deep RL methods. Lyapunov-based Actor-Critic \citep{han2020actor} integrates Lyapunov stability theory directly into the actor-critic framework. Here, the critic is trained to approximate a Lyapunov function, while the actor is optimised to select actions that decrease this function's value. This ensures that the policy not only maximises long-term reward but also drives the system toward a stable equilibrium. Similarly, Neural Lyapunov Control \citep{ya2019neural} learns the Lyapunov functions using neural networks, and constrains policy updates to satisfy Lyapunov decrease conditions. This allows stability guarantees to be maintained even in nonlinear and high-dimensional systems. Finally, Maximal Lyapunov Exponent Regularisation \citep{Young2024Enhancing} improves the stability of the control interaction by incorporating an estimation of local divergence in the policy loss for Dreamer V3 \citep{hafner2023dreamerv3}.

While these methods address stability in deep RL, they are limited in their scope as they rely on specific reward functions, specific model types or only constrain a subset of the system dynamics. In contrast, GIFT can be applied to any existing deep RL method and continuous control environment.

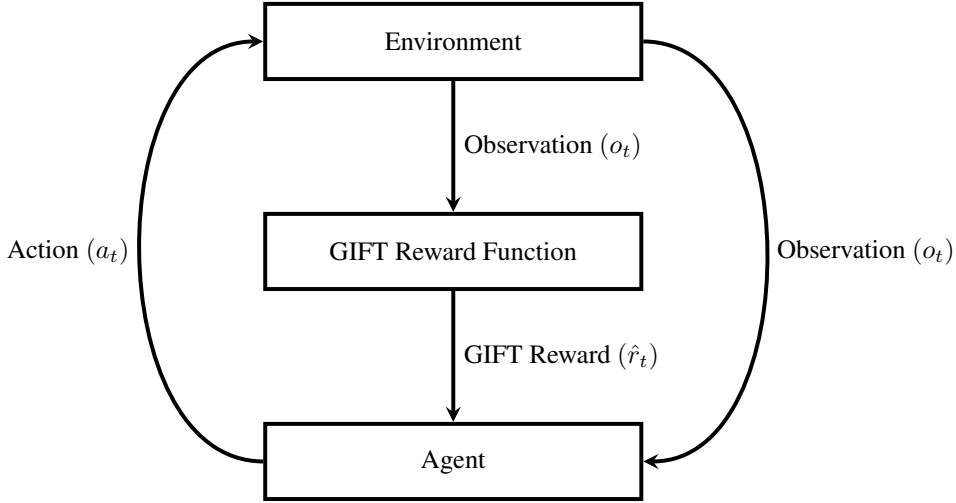
\begin{figure}[t]
    \centering
    \begin{tikzpicture}[node distance=.2\textwidth]
        \node (agent) [box, minimum width=5cm, minimum height=1cm] {Agent};
        \node (reward) [box, minimum width=5cm, minimum height=1cm, above of=agent] {GIFT Reward Function};
        \node (env) [box, minimum width=5cm, minimum height=1cm, above of=reward] {Environment};
        
        \draw [arrow] (env.east) to[bend left=90] node[anchor=west] {Observation $(o_t)$} (agent.east);
        \draw [arrow] (env.south) to[bend left=0] node[anchor=west] {Observation $(o_t)$} (reward.north);
        \draw [arrow] (reward.south) to[bend left=0] node[anchor=west] {GIFT Reward $(\hat{r}_t)$} (agent.north);
        \draw [arrow] (agent.west) to[bend left=90] node[anchor=east] {Action $(a_t)$} (env.west);
    \end{tikzpicture}
    
    \caption{Stabilising Markov Decision Processes}
    \label{fig:S-MDP}
\end{figure}

\newpage
\section{Methods} \label{sec:methods}
In this section, we present \textbf{G}lobal stabilisation via \textbf{I}ntrinsic \textbf{F}ine \textbf{T}uning (GIFT), a lightweight training framework designed to improve the stability of any deep Reinforcement Learning policy while preserving original task performance. This method is composed of three sequentially executed stages. First, a policy is trained to maximise the total reward in a given MDP. Second, a stabilising variation of the MDP is generated by replacing the standard reward function with a task-agnostic reward function which incentivises global stability. Finally, this policy is fine-tuned in the Stabilising MDP. These training stages are detailed in Sections~\ref{subsec:pre-training}-~\ref{subsec:fine-tuning}, and the whole algorithm is given in Algorithm~\ref{alg:GIFT}.

\subsection{Pre-training the baseline policy to maximise reward} \label{subsec:pre-training}
In the first stage of GIFT, a policy $(\pi_\theta)$ is trained in the MDP with the goal of maximising the total episodic reward $\left(\sum r_t\right)$. This training can be conducted using any deep RL method, and no modifications related to stability are introduced at this stage. The objective is to learn a policy that can achieve good performance in the original task. 

\subsection{Generating the Stabilising Markov Decision Process}
The second stage of GIFT constructs a \textit{Stabilising Markov Decision Process} (S-MDP) (Figure~\ref{fig:S-MDP}), a modified version of the original MDP designed to encourage globally stable behaviour. The S-MDP retains the same transition dynamics, observation space, and action space as the original MDP, but replaces the task-specific reward function with a task-agnostic intrinsic reward function $(\hat{R}:\mathcal{O} \times \mathcal{O} \rightarrow \mathbb{R})$ designed to promote convergence to a high-reward observation trajectory.

To define this intrinsic reward, GIFT leverages the pre-trained task-optimised policy to generate a dataset of $N \in \mathbb{N}$ observation and reward trajectories, each of length $L \in \mathbb{N}$. These trajectories are obtained by rolling out $\pi_\theta$ under varying initial conditions, and the observation trajectory that attains the highest total reward is selected as a \textit{reference trajectory} $(\mathrm{T}\in L \times \mathcal{O})$. This sequence defines a full rank region of the observation space associated with successful long-term performance, effectively defining optimal behaviour across all observation dimensions.

\newpage

The intrinsic reward function is designed to encourage the policy to take actions which not only manoeuvre the system to the high-performing region of the observation space but also track its temporal progression. Given a sample observation transition $(o_t, o_{t+1})$ and reference trajectory $\mathrm{T}$, GIFT first defines a target observation $(\tau(o_t))$ as the immediate successor of the observation in $\mathrm{T}$ that is closest to the current observation $o_t$: 

\begin{equation}
\label{eq:target_obs}
\tau(o_t) = \mathrm{T}\left[1 + \underset{l}{\arg\min}\left(||\mathrm{T}[l] - o_t||_2\right)\right].
\end{equation}

The intrinsic reward is then defined as:

\begin{equation}
\hat{R}(o_t, o_{t+1}) = \frac{1}{\left( \kappa \cdot ||o_{t+1} - \tau(o_t)||_2 \right)^2 + 1}.
\label{eq:GIFT_reward}
\end{equation}

The intrinsic reward $(\hat{r}_t = \hat{R}(o_t, o_{t+1}))$ is close to one when $o_{t+1}$ is near the target $\tau(o_t)$ and decays smoothly to zero as the distance between these observations increases, with the scalar $\kappa\in\mathbb{R}$ controlling the rate of this decay. This construction ensures that the policy is rewarded for remaining near the reference trajectory and progressing along it, thereby promoting convergence to both the spatial and temporal structure of the successful trajectory. Furthermore, as this function is defined over all dimensions of the observation space, it explicitly enforces global stability across all observable dimensions. 

\subsection{Fine-tuning to improve stability} \label{subsec:fine-tuning}
In the final stage of GIFT, the pre-trained policy is fine-tuned in the S-MDP to encourage globally stable behaviour. The objective of this stage is to optimise the policy to maximise the total intrinsic reward $\left(\sum \hat{r}_t\right)$ as defined in Equation~\ref{eq:GIFT_reward}. This fine-tuning is conducted by continuing to update the existing policy using any standard deep RL algorithm. Importantly, at this state, the task-specific reward is no longer used and the policy is now optimised solely with respect to the intrinsic reward $\hat{R}$. As a result, the fine-tuned policy is incentivised to suppress any unstable dynamics that were not penalised in the original MDP.


\begin{algorithm}[t]
\caption{\textbf{G}lobal stabilisation via \textbf{I}ntrinsic \textbf{F}ine \textbf{T}uning}
\label{alg:GIFT}
\begin{algorithmic}
\Require 
    \State Policy $(\pi_\theta)$
    \State Markov Decision Process (MDP)
    \State Learning Method (e.g. SAC / PPO)
\\
\Ensure $\pi_\theta$
\State \# Pre-training the baseline policy to maximise reward
\State Train $\pi_\theta$ in the MDP using the Learning Method
\State
\State \# Generating the Stabilising Markov Decision Process
\State Unroll the MDP for $L$ steps using $\pi_\theta$ starting in $N$ different inital positions
\State Set $\mathrm{T}$ as the observation trajectory which corresponds to the highest total reward
\State Generate the Stabilising MDP with the MDP dynamics and GIFT reward function (Equations~\ref{eq:target_obs}~\&~\ref{eq:GIFT_reward})
\State
\State \# Fine-tuning to improve stability
\State Fine-tune $\pi_\theta$ in the Stabilising MDP using the Learning Method
\State
\State \Return $\pi_\theta$
\end{algorithmic}
\end{algorithm}
\newpage
\section{Results} \label{sec:results}
This section evaluates the effectiveness of fine-tuning deep Reinforcement Learning policies using GIFT. The aim is to determine whether this additional training phase consistently enhances the stability of control interactions without compromising system performance. Section~\ref{subsec:experimental_setup} outlines our experimental setup, while Sections~\ref{subsec:performance} and \ref{subsec:stability} show how GIFT impacts performance and stability, respectively.

\subsection{Experimental Setup}\label{subsec:experimental_setup}
We apply GIFT to Soft Actor-Critic (SAC) \citep{haarnoja_2018_soft} and Proximal Policy Optimisation (PPO) \citep{Schulman2017Proximal} as these represent the state-of-the-art off-policy and on-policy deep RL methods. Each agent is trained in environments \textit{Walker} and \textit{Humanoid} environments sampled from the MuJoCo Playground \citep{zakka2025mujocoplayground} variation of the DeepMind Control Suite \citep{tassa_2018_deepmind}. These environments represent a range of complex continuous control tasks which are known to be unstable. 

Each policy-environment pair is independently trained with three initial seeds using an Intel Core i7-8700 CPU workstation, with an Nvidia RTX 2080 Ti GPU, and 32GB of RAM. After training the baseline policy, GIFT unrolls the MDP for $L=1000$ steps under the deterministic version of this policy, starting in $N=128$ unique initial states. From these, we select the observation trajectory associated with the highest cumulative reward to serve as the reference trajectory. Finally, we use the hyperparameters outlined in Appendix~\ref{app:hps} for both pre-training and fine-tuning and use $\kappa=\sqrt{17/3}$ when calculating the intrinsic reward.

\subsection{Performance}\label{subsec:performance}
To assess GIFT's impact on the performance of the control interaction, we measure the total reward for each policy-environment pair. We compute the total reward over 1000 steps starting from 100 distinct initial states. For each environment, we report the interquartile mean of the total reward in Table~\ref{tab:mle}, along with a bootstrapped 95\% confidence interval \cite{agarwal2021deep} to capture variability due to random seeds.  

\begin{figure}[t]
     \centering
    \begin{subfigure}[t]{\linewidth}
        \centering
    \includegraphics[width=\linewidth]{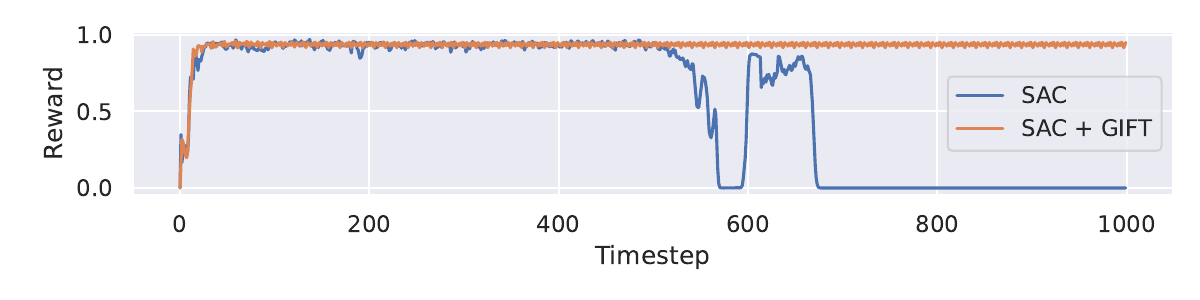}
        \caption{Reward Trajectories}
        \label{subfig:reward_trajectory_raw}
    \end{subfigure}%
    \\
    \begin{subfigure}[t]{\linewidth}
        \centering
        \includegraphics[width=\linewidth/9]{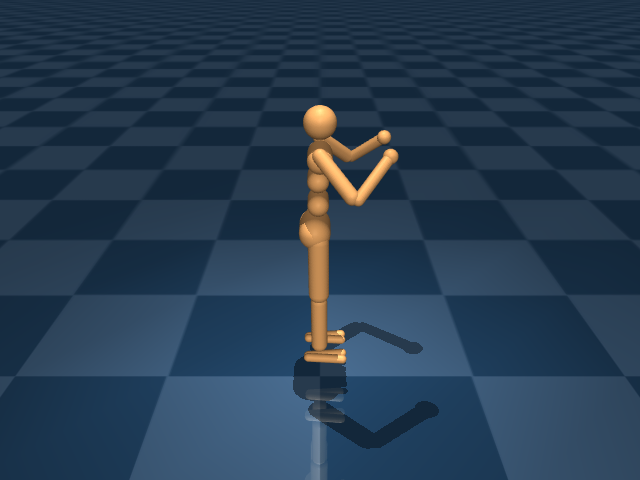}
        \includegraphics[width=\linewidth/9]{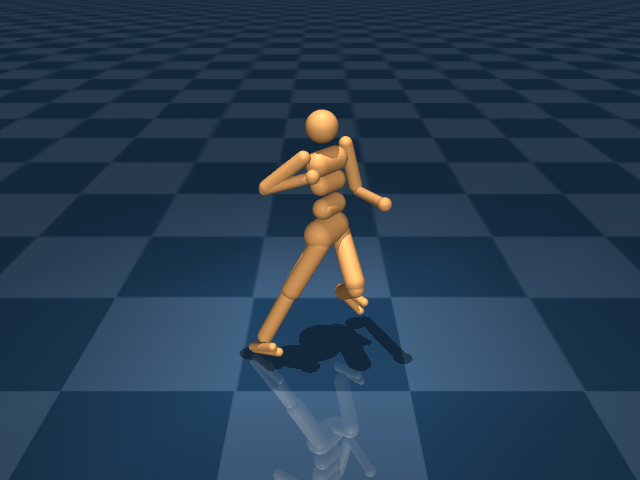}
        \includegraphics[width=\linewidth/9]{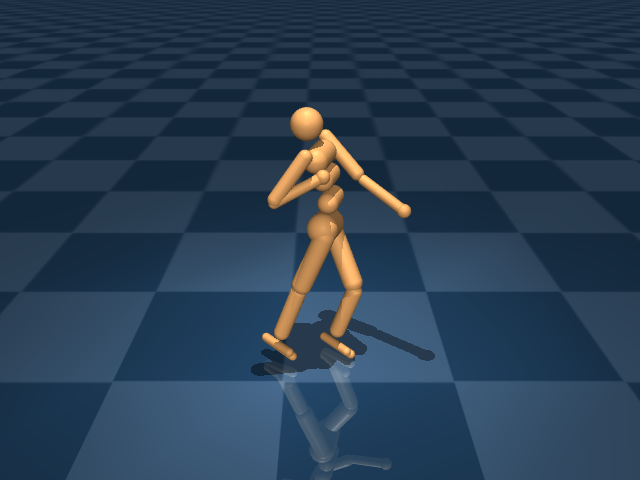}
        \includegraphics[width=\linewidth/9]{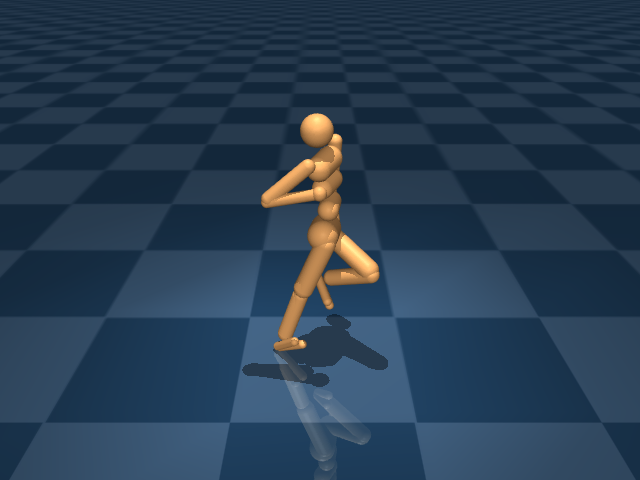}
        \includegraphics[width=\linewidth/9]{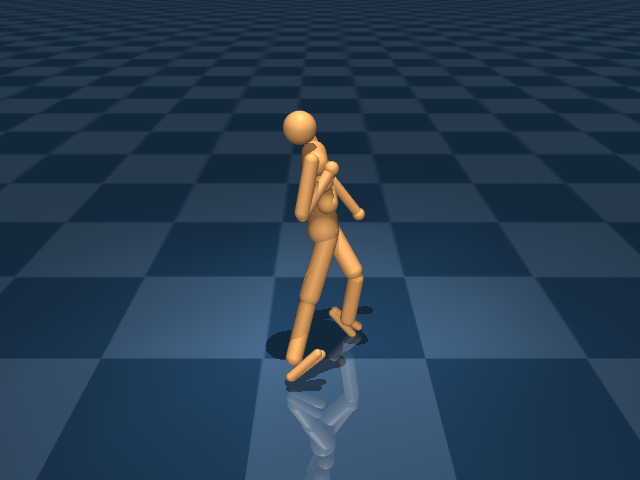}
        \includegraphics[width=\linewidth/9]{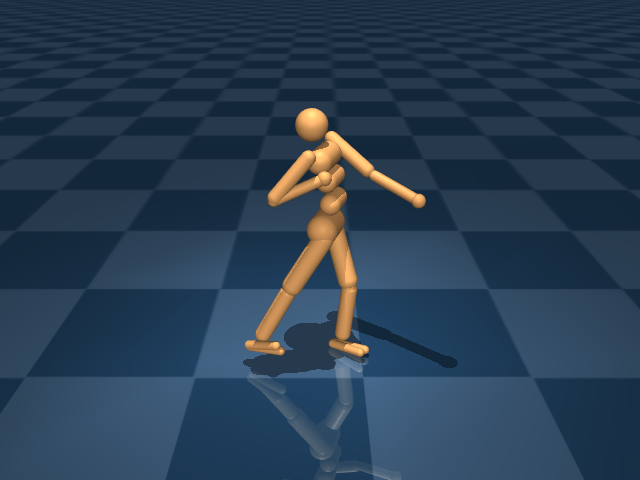}
        \includegraphics[width=\linewidth/9]{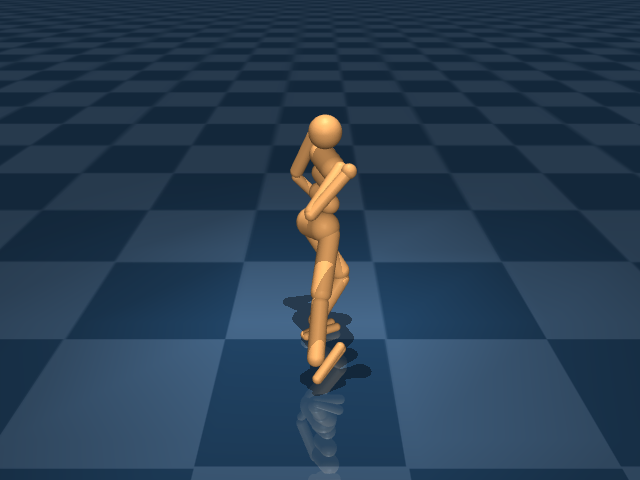}
        \includegraphics[width=\linewidth/9]{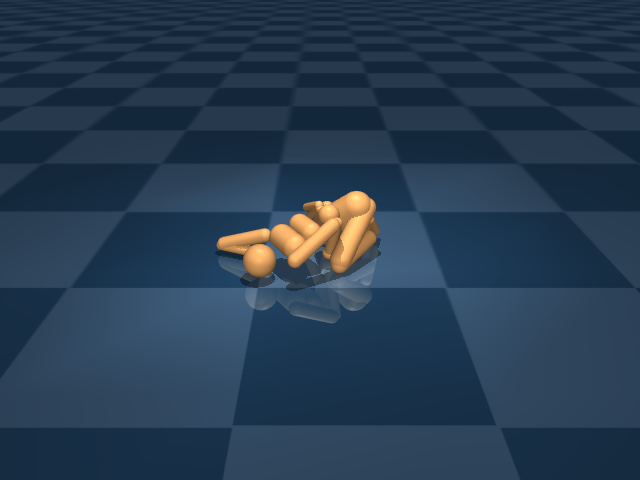}        
        \caption{SAC Render}
        \label{subfig:reward_trajectory_sac}
    \end{subfigure}%
    \\
    \begin{subfigure}[t]{\linewidth}
        \centering
        \includegraphics[width=\linewidth/9]{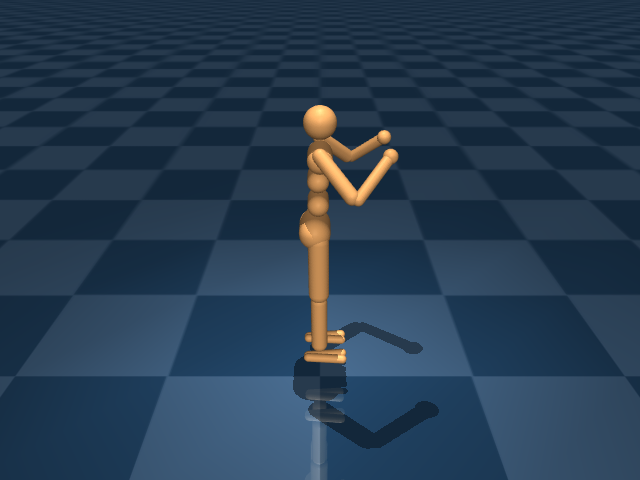}
        \includegraphics[width=\linewidth/9]{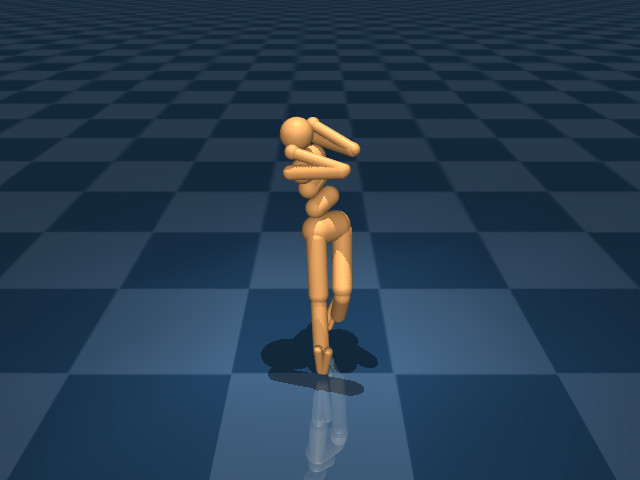}
        \includegraphics[width=\linewidth/9]{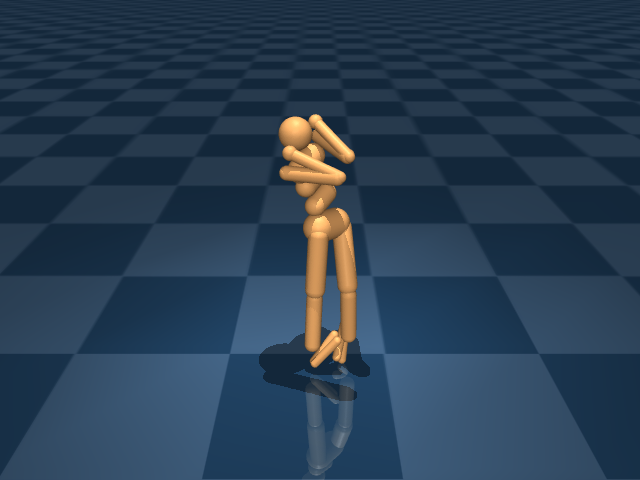}
        \includegraphics[width=\linewidth/9]{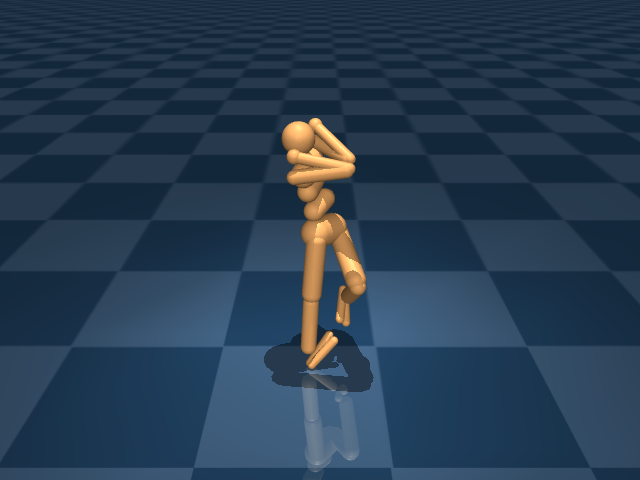}
        \includegraphics[width=\linewidth/9]{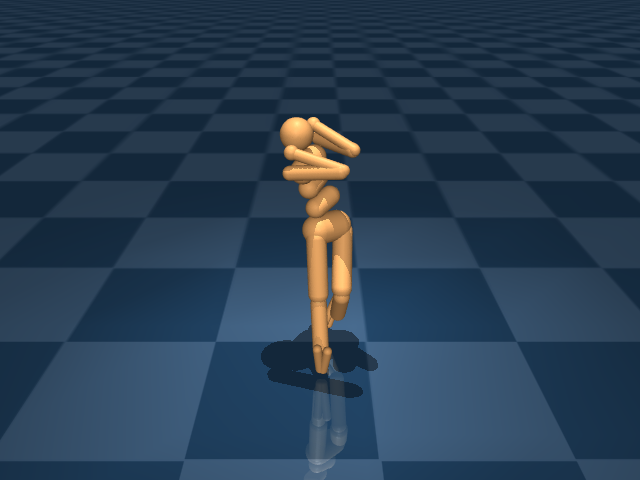}
        \includegraphics[width=\linewidth/9]{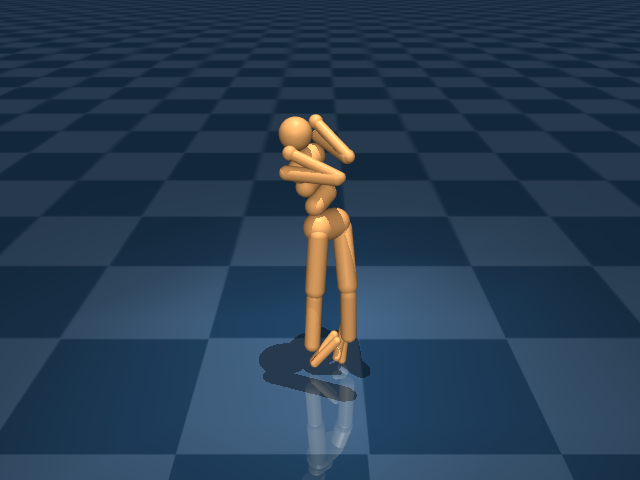}
        \includegraphics[width=\linewidth/9]{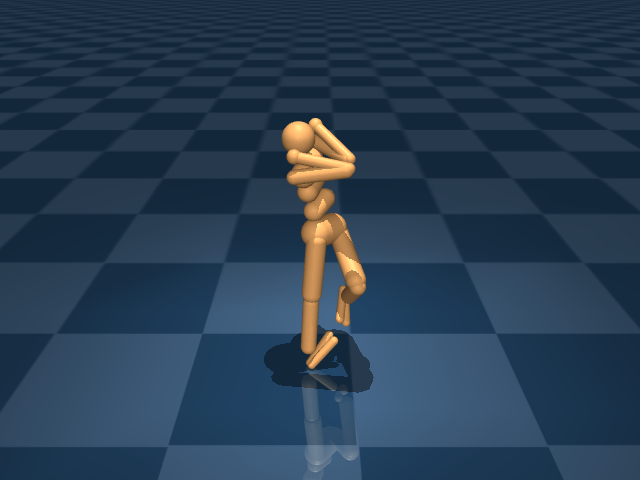}
        \includegraphics[width=\linewidth/9]{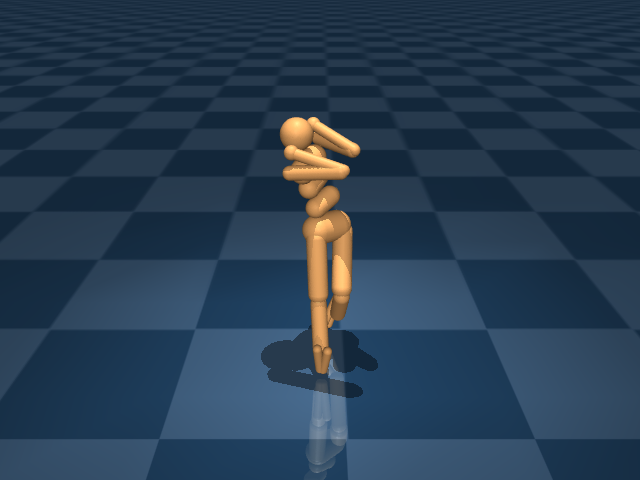}     
        \caption{SAC + GIFT Render}
        \label{subfig:reward_trajectory_sac_gift}
    \end{subfigure}%
    \caption{Reward trajectory attained by SAC and SAC + GIFT when controlling the \textit{Humanoid Walk} environment starting in the same initial position (Figure~\ref{subfig:reward_trajectory_raw}). SAC attains a lower total reward as the unstable system dynamics cause the robot to fall over (Figure~\ref{subfig:reward_trajectory_sac}). In contrast, when GIFT is applied to SAC, the robot remains vertical as the control interaction is more stable (Figure~\ref{subfig:reward_trajectory_sac_gift}).}
    \label{fig:reward_trajectory}
\end{figure}

These results indicate that GIFT does not adversely affect the performance of the control interaction. Across the majority of environments, incorporating GIFT preserves or modestly improves the total reward. This is particularly evident when SAC controls the \textit{Humanoid} environments, as policies fine-tuned with GIFT produce on average a 9\% increase in total reward. The performance increase occurs because the unstable dynamics produced by the standard SAC policy often cause the robot to fall over, thus yielding low rewards (Figure~\ref{fig:reward_trajectory}). Overall, these findings indicate that fine-tuning deep RL policies with GIFT can have a positive impact on the performance of the control interaction.

\begin{table}[t]
\caption{Average total reward produced by environments sampled from the \textit{DeepMind Control Suite} when controlled by trained deep RL policies. A bootstrapped 95\% confidence interval is included to show the variation introduced by initial seeds.  Higher values indicate a better-performing control interaction. \\}
\label{tab:total_reward}
\centering
\begin{tabular}{l|ll|ll}
\toprule
Environment         & SAC   & SAC + GIFT    & PPO   & PPO + GIFT         \\
\midrule
Walker Stand        & $985.3 \pm{~~1.2}$            & $\textbf{986.7} \pm{~~1.1}$   & $959.2 \pm{~~~~5.0}$              & $\textbf{969.0} \pm{~~~~3.2}$     \\
Walker Walk         & $\textbf{975.7} \pm{~~1.1}$   & $974.4 \pm{~~1.3}$            & $\textbf{960.5} \pm{~~~~3.9}$     & $903.8 \pm{~~41.8}$               \\
Walker Run          & $\textbf{758.8} \pm{~~5.5}$   & $735.1 \pm{26.0}$             & $\textbf{656.2} \pm{133.8}$       & $634.9 \pm{125.2}$                \\
Humanoid Stand      & $937.1 \pm{16.8}$             & $\textbf{961.7} \pm{11.6}$    & $~~\textbf{83.5} \pm{~~30.1}$     & $~~13.3 \pm{~~~~2.6}$             \\
Humanoid Walk       & $919.6 \pm{12.6}$             & $\textbf{939.7} \pm{10.2}$    & $~~\textbf{27.6} \pm{~~~~5.6}$    & $~~~~3.7 \pm{~~~~0.5}$            \\
Humanoid Run        & $282.1 \pm{14.2}$             & $\textbf{345.5} \pm{93.5}$    & $~~\textbf{11.8} \pm{~~~~2.1}$    & $~~~~7.6 \pm{~~~~4.8}$            \\
\bottomrule
\end{tabular}
\end{table}

\begin{table}[t]
\caption{Average Maximal Lyapunov Exponent produced by environments sampled from the \textit{DeepMind Control Suite} when controlled by trained deep RL policies. A bootstrapped 95\% confidence interval is included to show the variation introduced by initial seeds. Lower values indicate a more stable control interaction. \\}
\label{tab:mle}
\centering
\begin{tabular}{l|ll|ll}
\toprule
Environment         & SAC   & SAC + GIFT    & PPO   & PPO + GIFT         \\
\midrule
Walker Stand        & $0.0433 \pm{0.0318}$      & $\textbf{0.0036} \pm{0.0004}$      & $0.0084 \pm{0.0016}$             & $\textbf{0.0065} \pm{0.0012}$     \\
Walker Walk         & $0.0661 \pm{0.0170}$      & $\textbf{0.0047} \pm{0.0002}$      & $\textbf{0.0455} \pm{0.0213}$    & $0.0526 \pm{0.0194}$              \\
Walker Run          & $0.0841 \pm{0.0355}$      & $\textbf{0.0106} \pm{0.0015}$      & $0.1068 \pm{0.0024}$             & $\textbf{0.0635} \pm{0.0130}$     \\
Humanoid Stand      & $0.0654 \pm{0.0218}$      & $\textbf{0.0057} \pm{0.0021}$      & $0.0116 \pm{0.0024}$             & $\textbf{0.0036} \pm{0.0006}$     \\
Humanoid Walk       & $0.1918 \pm{0.0125}$      & $\textbf{0.0034} \pm{0.0003}$      & $0.0117 \pm{0.0037}$             & $\textbf{0.0048} \pm{0.0016}$     \\
Humanoid Run        & $0.2058 \pm{0.0371}$      & $\textbf{0.0040} \pm{0.0012}$      & $0.0096 \pm{0.0009}$             & $\textbf{0.0064} \pm{0.0037}$     \\
\bottomrule
\end{tabular}
\end{table}

\begin{figure}[t]
     \centering
    \begin{subfigure}[t]{0.4\linewidth}
        \centering
        \includegraphics[width=\linewidth]{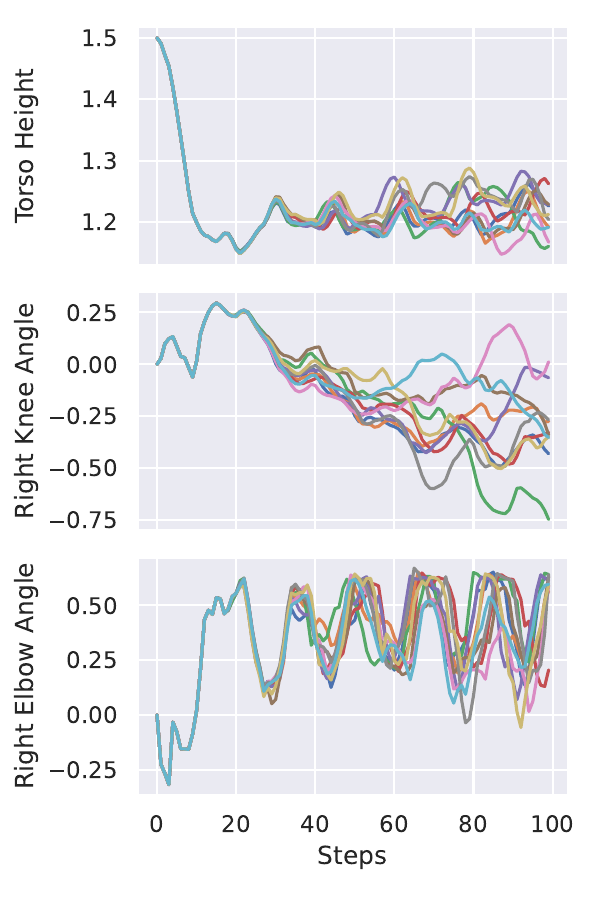}
        \caption{SAC}
        \label{subfig:dr3_cb_state_dynamics}
    \end{subfigure}%
    \hspace{0.08\linewidth}
    \begin{subfigure}[t]{0.4\linewidth}
        \centering
        \includegraphics[width=\linewidth]{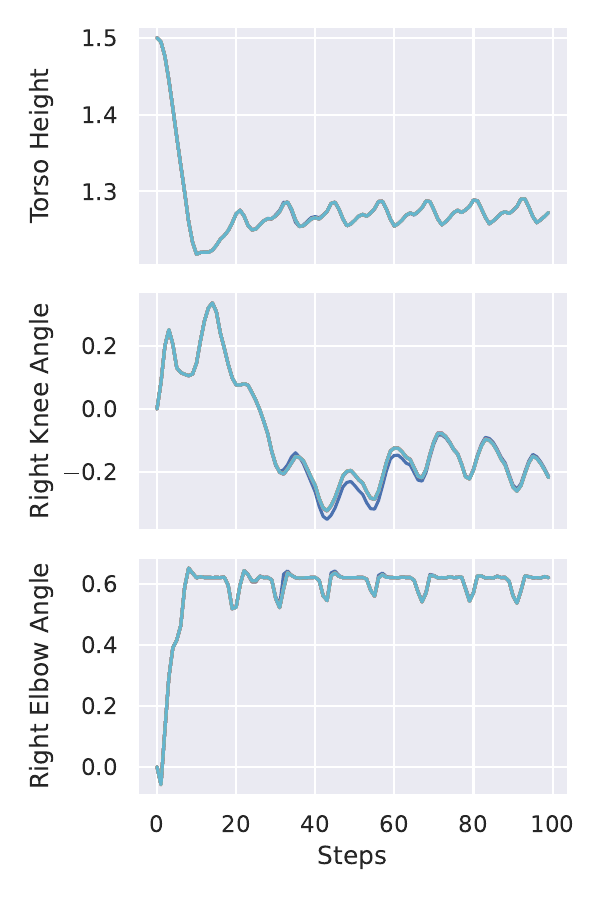}
        \caption{SAC + GIFT}
        \label{subfig:dr3_ww_state_dynamics}
    \end{subfigure}%
    \caption{Ten partial state trajectory produced by SAC (a) and SAC + GIFT (b) when controlling the \textit{Humanoid Walk} environment subject to a single initial state perturbation of size $10^{-4}$. The standard \textit{Humanoid Walk} reward function only constrains the torso height, angle and speed, while the GIFT reward function constrains all dimensions.}
    \label{fig:humanoid_walk_state_dynamcis}
\end{figure}

\subsection{Stability}\label{subsec:stability}
To assess GIFT's impact on the stability of the control interaction, we measure the Maximal Lyapunov Exponent (MLE) for each policy-environment pair. The MLE quantifies the exponential rate of divergence between state trajectories originating from nearby initial conditions. Positive MLE values indicate local instability, as small perturbations grow over time, while negative values indicate local convergence to an attractor. We compute the MLE using the Jacobian method \citep{cencini_2009_chaos, parker_2011_practical}, starting from 100 distinct initial states. For each environment, we report the interquartile mean MLE in Table~\ref{tab:mle}, along with a bootstrapped 95\% confidence interval to capture variability due to random seeds. These results demonstrate that GIFT substantially improves the stability of control policies across all systems. For both SAC and PPO, incorporating GIFT reduces the MLE by approximately an order of magnitude, indicating more consistent state trajectories under small state perturbations. 

\newpage

Furthermore, Figure~\ref{fig:humanoid_walk_state_dynamcis} shows a subset of the state trajectories produced by \textit{Humanoid Walk} when controlled by \textit{SAC} and \textit{SAC + GIFT} subject to an initial state perturbation of $10^{-4}$ units. This demonstrates that when maximising reward alone, SAC induces chaotic state dynamics, as this small change in system state compounds over time to produce substantially different outcomes long-term.  In contrast, the SAC + GIFT policy remains stable, as the perturbed trajectories remain close over time.
\section{Limitations} \label{sec:limitations}
GIFT is a general-purpose framework that improves the stability of existing deep RL policies by fine-tuning them using a custom reward function that encourages convergence towards a high-performing target trajectory. However, selecting the observation trajectory with the highest total reward as the stabilisation target can introduce undesirable effects. If the initial conditions deviate significantly from those seen during the generation of this trajectory, convergence may be poor or unstable. In addition, the selected target trajectory may lie in a region of the state space that is inherently unstable or difficult to influence due to limited controllability. In such cases, even a well-tuned stabilising reward may be insufficient to ensure reliable convergence. Finally, because GIFT operates in the observation space, its effectiveness may be limited in partially observable Markov decision processes. In such settings, the observed state may not fully capture the underlying dynamics, making it difficult to identify meaningful stabilisation targets or ensure consistent convergence.

\newpage
\section{Conclusion}
This paper introduced \textbf{G}lobal stabilisation via \textbf{I}ntrinsic \textbf{F}ine \textbf{T}uning (GIFT), a general-purpose framework for improving the global stability of deep reinforcement learning (RL) policies. GIFT operates by fine-tuning pre-trained policies within a Stabilising Markov Decision Process (S-MDP), which shares the same dynamics and observation space as the original task but employs an intrinsically generated reward function that penalises divergence in all observation dimensions. This explicit emphasis on constraining the full observation space enables GIFT to suppress chaotic dynamics that arise from underspecified reward signals. Empirical results show that GIFT reduces the Maximal Lyapunov Exponent by an order of magnitude while preserving task performance, demonstrating its potential to enhance the stability of deep RL in real-world continuous control environments.




\bibliographystyle{plainnat}
\bibliography{neurips_2025}

\newpage
\appendix

\section{Appendix}

\subsection{Hyperparameters}\label{app:hps}
\begin{table}[h]
    \centering
\caption{Hyperparameters used to train SAC}
    \begin{tabular}{lr}
    \toprule
    Total Timesteps & $5\times10^6$ \\
    Reward Scaling & $1.0$ \\
    Normalize Observations & True \\
    Discounting & $0.99$ \\
    Learning Rate & $10^{-3}$ \\
    \# Environments  & $128$ \\
    Batch Size & $512$ \\
    \# Gradient Updates Per Step & $8$ \\
    Maximum Replay Buffer Size & $2^{22}$ \\
    \bottomrule
    \end{tabular}
\end{table}

\begin{table}[h]
    \centering
\caption{Hyperparameters used to train PPO}
    \begin{tabular}{lr}
    \toprule
    Parameter & Value \\
    \midrule
    Total Timesteps & $6\times10^7$ \\ 
    Reward Scaling & $10.0$ \\ 
    Mormalize Observations & True \\ 
    Unroll Length & $30$ \\ 
    \# Minibatches & $32$ \\ 
    \# Updates Per Batch & $16$ \\ 
    Discounting & 0.995 \\ 
    Learning Rate & $10^{-3}$ \\ 
    Entropy Cost & $10^{-3}$ \\ 
    \# Environments & $2048$ \\ 
    Batch Size & $1024$ \\ 
    \bottomrule
    \end{tabular}
\end{table}


\end{document}